\documentclass{article}

\usepackage[final]{nips_2017}

\usepackage[utf8]{inputenc} 
\usepackage[T1]{fontenc}    
\usepackage{hyperref}       
\usepackage{url}            
\usepackage{booktabs}       
\usepackage{amsfonts}       
\usepackage{nicefrac}       
\usepackage{microtype}      
\usepackage{amsmath}
\usepackage{algorithm}
\usepackage[noend]{algpseudocode}
\usepackage{graphicx}
\usepackage{color,soul}
\usepackage{array}
\usepackage{tabu}
\usepackage{caption}
\usepackage{float}

\title{Asynchronous Advantage Actor-Critic Agent for Starcraft II}

\author{
  Basel Alghanem\\
  School of Computer Science\\
  Carnegie Mellon University\\
  Pittsburgh, PA 15213 \\
  \texttt{basela@cmu.edu} \\
  \And
  Keerthana P G \\
  School of Computer Science\\
  Carnegie Mellon University\\
  Pittsburgh, PA 15213 \\
  \texttt{kgopalak@andrew.cmu.edu} \\
}

\begin{document}

\maketitle

\begin{abstract}
  Deep reinforcement learning, and especially the Asynchronous Advantage Actor-Critic algorithm, has been successfully used to achieve super-human performance in a variety of video games. Starcraft II is a new challenge for the reinforcement learning community with the release of pysc2 learning environment proposed by Google Deepmind and Blizzard Entertainment. Despite being a target for several AI developers, few have achieved human level performance. In this project we explain the complexities of this environment and discuss the results from our experiments on the environment. We have compared various architectures and have proved that transfer learning can be an effective paradigm in reinforcement learning research for complex scenarios requiring skill transfer. 
\end{abstract}

\section{Introduction}
\label{intro}
Resurgence in deep reinforcement learning[10] research has resulted in the development of robust algorithms that can deliver human level performance in domains such as Atari [1], the game of Go [2], three-dimensional virtual environments [3] and simulated robotics domains [4, 5]. Many of these recent successes have been stimulated by the availability of simulated domains with an appropriate level of difficulty. To go beyond the benchmarks, the research community is required to develop algorithms in domains that are beyond the capabilities of current methods.

SC2LE (StarCraft II Learning Environment)[6] is a new challenge for exploring reinforcement learning algorithms and architectures, based on the StarCraft II video game. StarCraft is a real-time strategy (RTS) game that combines fast paced micro-actions with the need for high-level planning and execution. The game has been an enduring sport among humans for over 2 decades with millions of casual and highly competitive professional players, therefore, defeating top human players becomes a meaningful long-term objective.

\subsection{Related Work}

Mnih, et. al. [9] first proposed the Asynchronous Advantage Actor-Critic (A3C) algorithm, and applied it to Atari video games. Atari games have similar input state space to Starcraft II, in that the input is screens. However, the underlying state of Starcraft II is much more complicated because it has many more layers of information. Additionally, the action space of Starcraft is orders of magnitude larger. These differences are discussed further in \ref{env}.

The original paper proposing Starcraft as a new deep RL challenge by Vinyals et. al. [6] provides important benchmarks for our project. They used A3C with three different network architectures to produce results across several minigames as well as on one map of the full game.

Andersen, et. al., [17] applied deep reinforcement learning to Tower Line Wars, a game with complexity between Atari games and Starcraft II. They were ultimately unable to outperform simple rule-based agents. They observed that it is difficult for a deep network to develop an accurate state representation for a game with spatial actions.

Certicky \& Churchill [15] discussed the current state of Starcraft II AI competitions. Most of the bots discussed have substantial human-designed architecture. For example, they use different approaches for deciding which task to perform next or how to target opposing units. Justesen \& Risi [16] discuss one such approach for macromanagement. They collected state-action pairs of expert players and were able to predict the next build task of the experts 21\% of the time. In contrast, a deep reinforcement learning approach would only use a human-designed agent that would learn to make large and small decisions.

In a class project similar to this one, Chang [13] used A3C and a network architecture similar to the baseline network proposed by Vinyals et. al. [6]. He noted that network convergence depended strongly on hyperparameter selection.

In another class project, Barratt \& Pan [14] attempted to use imitation learning, using data collected from an experienced Starcraft II player, to teach an agent advanced tactics. While their agent was able to achieve the basic goal of the minigame they selected, it was not able to execute the complex strategies of the expert  player because their number of training samples was small compared to the great size of the state and action spaces.

\subsection{Project Goals}

We had a few goals for this project. We sought to determine what network architectures will allow the agent to converge quickly and consistently. Additionally, we considered the trade-off between convergence speed and whether the agent has converged on a suboptimal policy. Finally, we investigated how well we can transfer learning to different minigames. For example, what improvements in learning can be achieved by training an agent in one scenario which was pre-trained on a different scenario?

\section{Methods}
\label{methods}
In this project we used the A3C algorithm to yield results on the Starcraft 2 learning environment.

\subsection{Environment}
\label{env}
The SC2 environment is different from prior work in RL in many ways. First, it is a multi-agent problem in which several players compete for influence and resources. Second, it is an imperfect information game. The map is only partially observed via a local camera, which must be actively moved in order for the player to integrate information. Furthermore, it is necessary to actively explore the map in order to determine the opponent’s state. Third, the action space is vast and diverse with many different unit and building types, each with unique local actions. Furthermore, the set of legal actions available varies as the player progresses. Fourth, it has delayed credit assignment requiring long-term strategies over thousands of steps. 

We interact with the environment using PySC2, an open source python wrapper optimised for RL agents. PySC2 defines an action and observation specification to ease the interaction between Python reinforcement learning agents and StarCraft II. It also includes some minigames as challenges and visualisation tools to understand what the agent can see and do.

\textbf{Reward structure}\\
The agent plays against an in-built bot and to win a game, it must: 1. Accumulate resources, 2. Construct production buildings, 3. Amass an army, and 4. Eliminate all of the opponent’s buildings. The built-in AI is rule-based and comes with 10 difficulty levels. There are two different reward structures: ternary(1(win), 0(tie), -1(loss)) received at the end of a game and  a Blizzard score. The Blizzard score increases with more mined resources, decreases when losing units/buildings, and is not affected by other actions (training units, building buildings, and researching). The Blizzard score is far less sparse than the ternary reward signal and can act as continual feedback for the agent. \\\\
\textbf{Observations}\\
The main observations come as sets of feature layers which depict specific information such as unit types, hit points, visibility, etc which are rendered at N*M pixels (where N and M are configurable). There are two sets of feature layers: the minimap is a coarse representation of the state of the entire world, and the screen is a detailed view of a subsection of the world corresponding to the player’s on-screen view. There are 13 screens and 7 minimaps. In addition, there are various non-spatial observations such as the amount of gas and minerals collected, the set of actions currently available, information about selected units, build queues, and units in a transport vehicle, etc. \\\\
\textbf{Actions}\\
In StarCraft, not all actions are available in every game state. For example, the move command is only available if a unit is selected. The number arguments required to call each action are also different. Here, an action $a$ is represented as a composition of a function identifier a0 and a sequence of arguments which that function identifier requires: $a_1, a_2, . . . , a_L$. To represent the full action space, there are 524 action-function identifiers with 13 possible types of arguments. The response rate of the agent can also be configured (default at one action every 8 game frames/ 180 actions per minute (APM). Notably, human players take between 30 and 500 APM.

\subsection{A3C Algorithm}

The Asynchronous Advantage Actor-Critic (A3C) algorithm (\ref{algo}) was proposed by Mnih, et. al. [9]. It builds on top of the Actor-Critic method, where two functions are developed: one that maintains and updates a policy (the actor) and another that evaluates the current state (the critic). In Advantage Actor-Critic (A2C), the critic's evaluation uses an advantage function, rather than a value function. The actor's policy is updated according to the critic's advantage function, and the critic's advantage function is updated according to transitions generated by the actor's interaction with the environment.

\begin{algorithm}
\caption{Asynchronous advantage actor-critic - pseudocode for each actor-learner thread.}\label{alg:a3c}
\begin{algorithmic}
\State \textit{// Assume global shared parameter vectors $\theta$ and $\theta_v$ and global shared counter $T = 0$}
\State \textit{// Assume thread-specific parameter vector $\theta'$ and $\theta'_{v}$}
\State Initialize thread step counter $ t \leftarrow 1 $
\Repeat 
\State Reset gradients: $ d\theta \leftarrow 0$ and $d \theta_{v} \leftarrow 0 $
\State Synchronize thread-specific parameters $\theta'=\theta$ and $\theta'_{v} = \theta_v$
\State $t_{start} = t$
\State Get state $s_t$
\Repeat
\State Perform $a_t$ according to policy $\pi(a_t|s_t;\theta')$
\State Receive reward $r_t$ and new state $s_{t+1}$
\State $t \leftarrow t + 1$
\State $T \leftarrow T + 1$
\Until{terminal $s_t $ \textbf{or} $ t-t_{start}==t_{max} $}
\State $R = \begin{cases} 
      0 & $ for terminal $ s_t \\
      V(s_t,\theta'_{v}) & $ for terminal $ s_t $ \textit{ // Bootstrap from last state} $ \\
   \end{cases}
$
\For{$i \in {t-1,\dots,t_{start}}$}
\State $R \leftarrow r_i + \gamma R$
\State Accumulate gradients wrt $\theta'$: $d \theta \leftarrow d \theta + \nabla_{\theta'}\log \pi (a_i|s_i;\theta')(R-V(s_i;\theta'_v))$
\State Accumulate gradients wrt $\theta'_v$: $ d \theta_v + \partial(R-V(s_i;\theta'_v))^2/\partial \theta'_v $
\EndFor
\State Perform asynchronous update of $\theta$ using $d \theta $ and of $\theta_v$ using $d \theta_v$
\Until{$T>T_{max}$}
\end{algorithmic}
\label{algo}
\end{algorithm}

The A3C algorithm has multiple agents executing in parallel. Each agent regularly learns, passes on that learning to shared memory, then reloads to that shared memory. These asynchronous updates are an alternative to experience replay and was first proposed by Nair, et. al. [11]. Experience replay aggregates memory and reduces non-stationarity and decorrelated updates in reinforcement learning. However, using experience replay limits the methods to off-policy algorithms that can update from data generated by an older policy. Moreover, it uses more memory and incurs more computation per real interaction. The parallelism of asynchronous updates decorrelates the data since at any given time, the parallel agents experience a variety of states. This allows the execution of on-policy algorithms like Sarsa and actor-critic. Additionally, asynchronous agents can be run on a standard multicore CPU instead of using specialized hardware.

A3C also improves the diversity of the learning experience as multiple actors-learners running in parallel are likely to be exploring different parts of the environment. This diversity can be maximised by explicitly using different exploration policies and native behaviours in each actor-learner. Using multiple parallel actor learnings is reported to have resulted in a reduction in training time that is roughly linear in the number of parallel actor-learners[9]. 

\subsection{Proposed Framework \& Challenges}
\begin{figure}[!h]
  \centering
  \includegraphics[width=70mm]{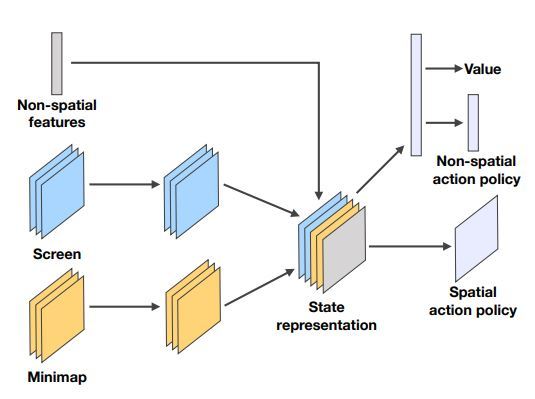}
  \caption{Network Architecture[6]}
  \label{framework}
\end{figure}
There are several challenges here. Previous work using A3C in Atari games have either used fully spatial features such as images or fully non-spatial features, but here, the agent requires both classes of information to play well. Additionally, the action space is very large and is a function of the current state. Actions fall under two categories: spatial and non-spatial. Spatial actions require a spatial argument, which is a pixel on the game map where the action should be taken. For each spatial action, there are over 7,000 underlying actions. The combination of large state space and large action space was a major challenge.

The framework we used is shown in \ref{framework}. Our software was adapted from a pre-existing library developed by Xiaowei Hu [12] which implemented a network designed by Vinyals, et. al. [6]. To account for spatial and non-spatial actions, different policy outputs are included. The non-spatial action output indicates which action should be taken and the spatial action output indicates where actions that are spatial should be taken. This means that, for the correct spatial arguments to be selected, the state representation should be enough information for the part of the network specific to spatial actions to guess what the best action will be.

We used three variations on this one network, described in \ref{networks}. The first was the baseline FullyConv (hereafter referred to as "Baseline") agent proposed by Vinyals et. al. [6]. The second, PlusFC, used an additional fully connected layer before each of the three network outputs (spatial action, non-spatial action, and value). The third used an additional convolutional layer specific to the minimaps and the screens before they're concatenated. All three networks used the same parameters, described in \ref{params}.

\begin{figure}[!h]
\centering
\captionof{table}{Network Architectures Used} 
\begin{tabular}{ p{4.5cm} | p{25mm}| p{25mm}| p{25mm}} 
 & Baseline & PlusFC & PlusConv  \\ 
\hline
Minimap Input Processing & 16 ch. 5x5 conv \newline 32 ch. 3x3 conv& 16 ch. 5x5 conv \newline 32 ch. 3x3 conv & 32 ch. 5x5 conv \newline 48 ch. 3x3 conv \newline 16 ch. 3x3 conv  \\ 
\hline
Screen Input Processing & 16 ch. 5x5 conv \newline 32 ch. 3x3 conv & 16 ch. 5x5 conv \newline 32 ch. 3x3 conv & 32 ch. 5x5 conv \newline 48 ch. 3x3 conv \newline 16 ch. 3x3 conv   \\ 
\hline
Flat Inputs Processing & 256-unit FC layer & 256-unit FC layer & 256-unit FC layer  \\ 
\hline
Value Output Processing & 1 shared 256-unit FC layer& 1 shared 256-unit FC layer \newline 128-unit FC layer & 1 shared 256-unit FC layer  \\ 
\hline
Non-Spatial Output Processing & 1 shared 256-unit FC layer& 1 shared 256-unit FC layer \newline 128-unit FC layer & 1 shared 256-unit FC layer  \\ 
\hline
Spatial Output Processing & 1 ch. 1x1 conv & 1 ch. 1x1 conv & 1 ch. 1x1 conv  \\ 
\end{tabular}
\label{networks}
\end{figure}

\begin{figure}[H]
\centering
\captionof{table}{Training Parameters Used} 
\begin{tabular}{ | p{12em} | p{5.5cm}| } 
\hline
Parameters& Value \\ 
\hline
Learning Rate& $5*10^{-4}$ \\ 
\hline
Discount Factor & 0.99 \\ 
\hline
Screen resolution & 64x64 \\ 
\hline
Minimap resolution & 64x64 \\ 
\hline
No. of asynchronous agents & 4, 8, or 16 based on machine capacity \\ 
\hline
Frequency of agent's action & 1 per 8 steps of sc2bot \\ 
\hline
Exploration strategy & Epsilon greedy \\ 
\hline
\end{tabular}
\label{params}
\end{figure}

\section{Results}
\label{results}

First, we have collected benchmarks to put our progress in context. For the minigame scenarios, we can compare against the leaderboard at \url{starcraftgym.com/}. We can also compare against a random agent playing the same minigames to see whether the agent has begun to converge. The maximum score achieved by each network architecture trained on each scenario is shown in \ref{maxvals}

\begin{figure}[!h]
\centering
\captionof{table}{Maximum Scores obtained} 
\begin{tabular}{ | m{12em} | m{2cm}| m{2cm} | m{2cm} |} 
 \hline
 Map & Baseline & PlusFC & PlusConv \\
 \hline
 \hline
 MoveToBeacon  & 29  &-  &-   \\
\hline 
 CollectMineralShards  & 33  & - & 73  \\
\hline
 DefeatRoaches  & 42  & - & 54  \\
\hline
 FindAndDefeatZerglings  & 16  & 9(in 16k updates) & -  \\
\hline
DefeatBanelingsAndZerglings & 113(with transfer in 4k episodes) & 130  & -  \\
\hline
BuildMarines  & 0  & - & -  \\
\hline
Simple64  & 6750  & - & -  \\
\hline
\end{tabular}
\label{maxvals}
\end{figure}

\subsection{Map By Map Discussion}
\begin{enumerate}
  \item MoveToBeacon: Here, the agent earns rewards by moving the single unit to a beacon. Whenever the agent earns a reward for reaching the beacon, the beacon is teleported to a random location.

Using the baseline agent, we are able to perform very well and get human-level performance. Our performance is similar to the top performances on the leaderboard. Video of trained agent can be viewed \href{https://www.youtube.com/watch?v=26TbPYKrPco&feature=youtu.be}{here}. The results for this scenario are shown in \ref{MTB}.

\begin{figure}[!h]
  \centering
  \includegraphics[width=80mm]{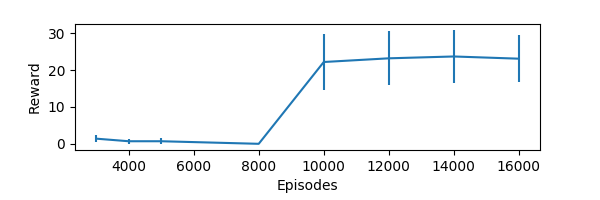}
  \caption{MoveToBeacon Rewards}
  \label{MTB}
\end{figure}

  \item DefeatRoaches: Here, the objective is to defeat units labeled Roaches. Video of trained agent can be viewed \href{https://www.youtube.com/watch?v=2uT2nkilCng}{here}. We were able to achieve close to the optimal behavior, but we were unable to achieve leaderboard-level results. The results for this scenario are shown in \ref{DR}.
  
  \begin{figure}[!h]
  \centering
  \includegraphics[width=80mm]{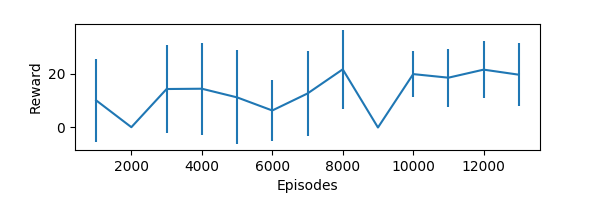}
  \caption{DefeatRoaches Rewards}
  \label{DR}
\end{figure}

  \item CollectMineralShards: Here, the objective is to use 2 marines to collect minerals in the visible section of map. While the optimal policy is to send the marines in two different directions (thereby collecting minerals twice as fast), our agent learns a successful but sub-optimal policy where the two marines collect minerals together. Video of trained agent can be viewed \href{https://www.youtube.com/watch?v=6TnkOidvieM}{here}. 
  
  For this scenario, we tested transfer learning. We took the final model weights from our DefeatRoaches network and used it as the starting point for PlusConv training. Both scenarios involve selecting units and moving them within the original screen. As shown in \ref{CollectMineralShards}, this transfer learning resulted in substantially better results than training the Baseline network and substantially faster convergence than training the PlusConv network from scratch. We were not able to achieve leaderboard-level results.
  
  \begin{figure}[!h]
  \centering
  \includegraphics[width=0.55\textwidth]{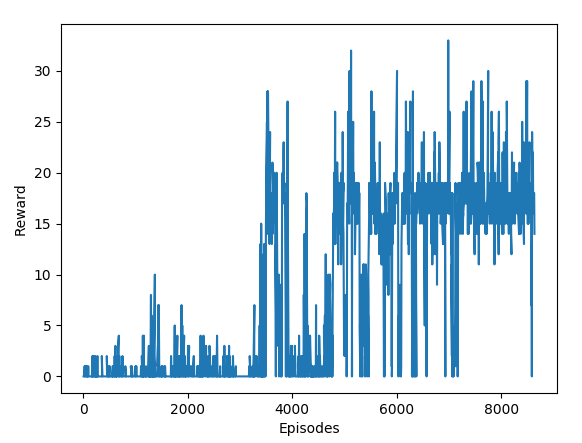}
  \includegraphics[width=0.45\textwidth]{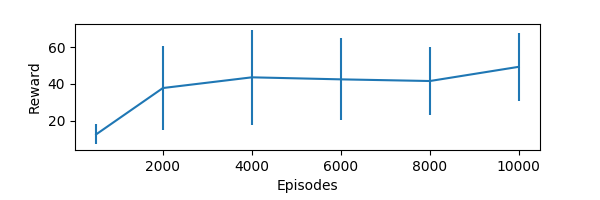}
  \includegraphics[width=0.45\textwidth]{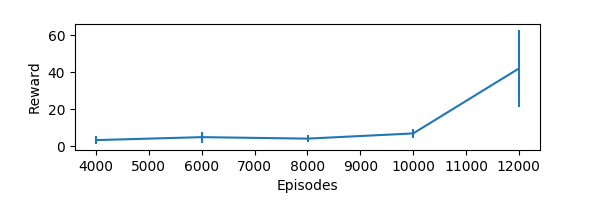}
  \caption{Baseline (top) vs PlusConv with Transfer vs PlusConv from scratch for the CollectMineralShards minigame}
  \label{CollectMineralShards}
\end{figure}

  \item FindAndDefeatZerglings: Here, the agent needs to balance exploration and combat to find stationary Zerglings on the full-map and destroy them. After 97k episodes, our agent is good at defeating Zerglings that are already found but doesn’t explore the entire map to find all the Zerglings. About 30-40\% of the map is left unexplored. It can further be seen that during the initial training, subsequent models improve on their exploration capability. Longer training, with forcing exploration may improve performance. Video of trained agent at 97k can be viewed \href{https://www.youtube.com/watch?v=OpK1zUE7xnw&feature=youtu.be}{here}. Results from this scenario are shown in \ref{FADZ}
  \begin{figure}[!h]
  \centering
  \includegraphics[width=0.5\textwidth]{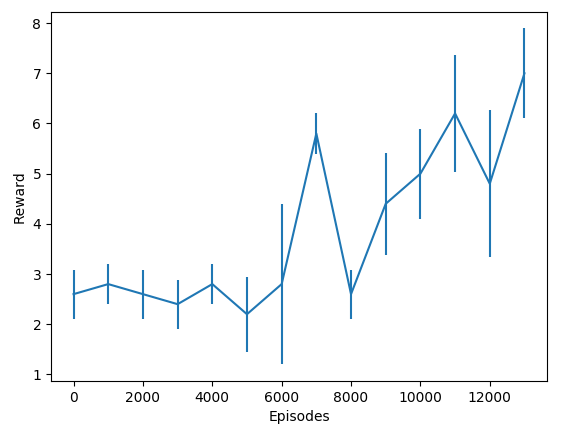}  
  \caption{Baseline and PlusFC for the FindAndDefeatZerglings minigame}
  \label{FADZ}
\end{figure}

  \item DefeatZerglingsAndBanelings: Agent needs to control and move Marines to destroy Zerglings and Banelings which are enemy units. Video of deep-FCN trained agent, without transfer learning can be viewed \href{https://www.youtube.com/watch?v=eCn8ioShf5M&feature=youtu.be}{here}. We also took the network learned from FindAndDefeatZerglings and used that as a starting point to train on DefeatZerglingsAndBanelings. These results are shown in \ref{DZAB}. An interesting observation is that FindAndDefeatZerglings learns a policy of moving all marines together as a unit. While the agent at first does well using this policy on the new map, scores begin to drop as it explores newer policies where the marines split up to attack and eventually learns a more optimal policy and scores improve.
  \begin{figure}[h]
  \centering
  \includegraphics[width=0.6\textwidth]{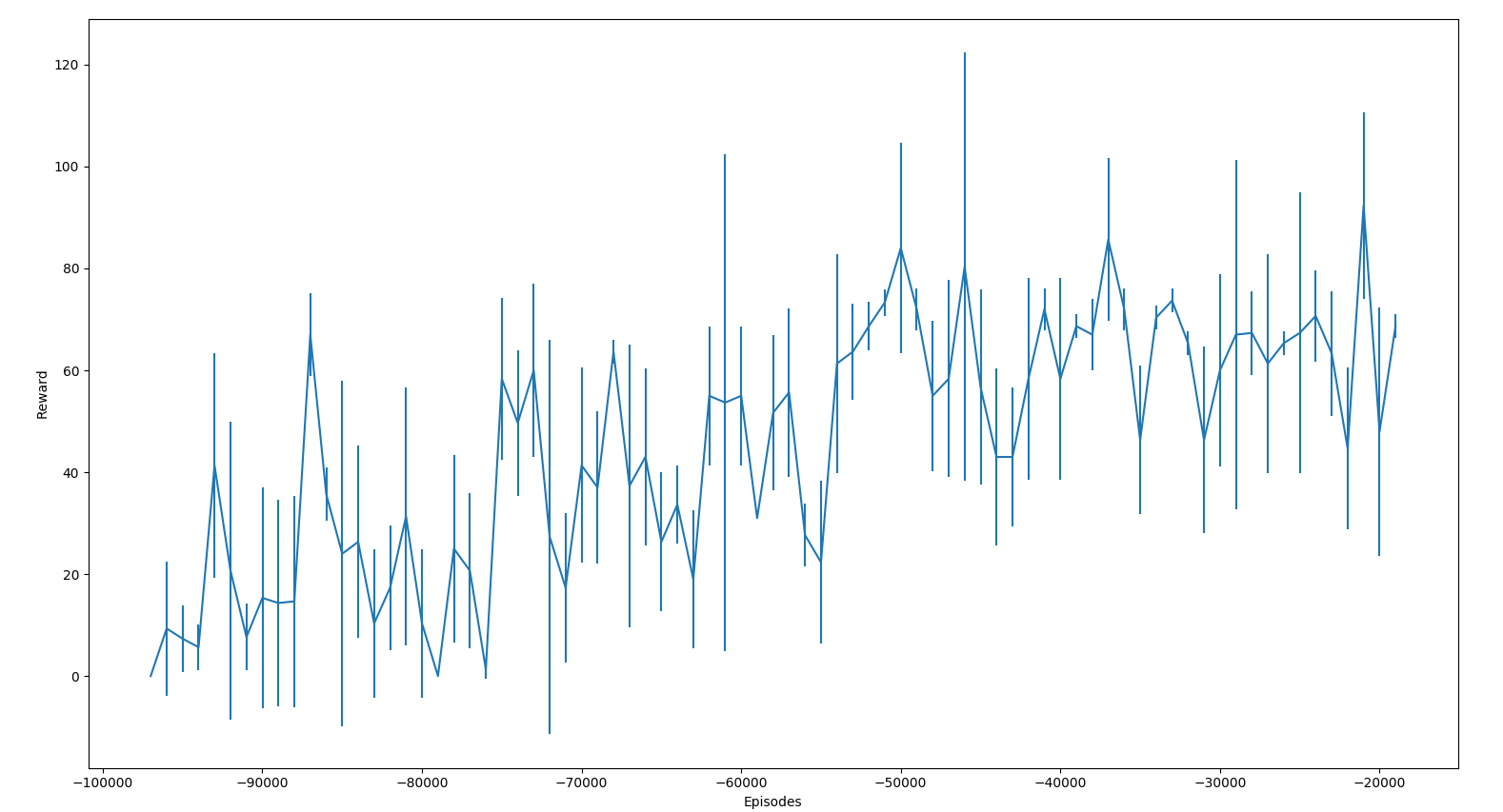}
  \includegraphics[width=0.6\textwidth]{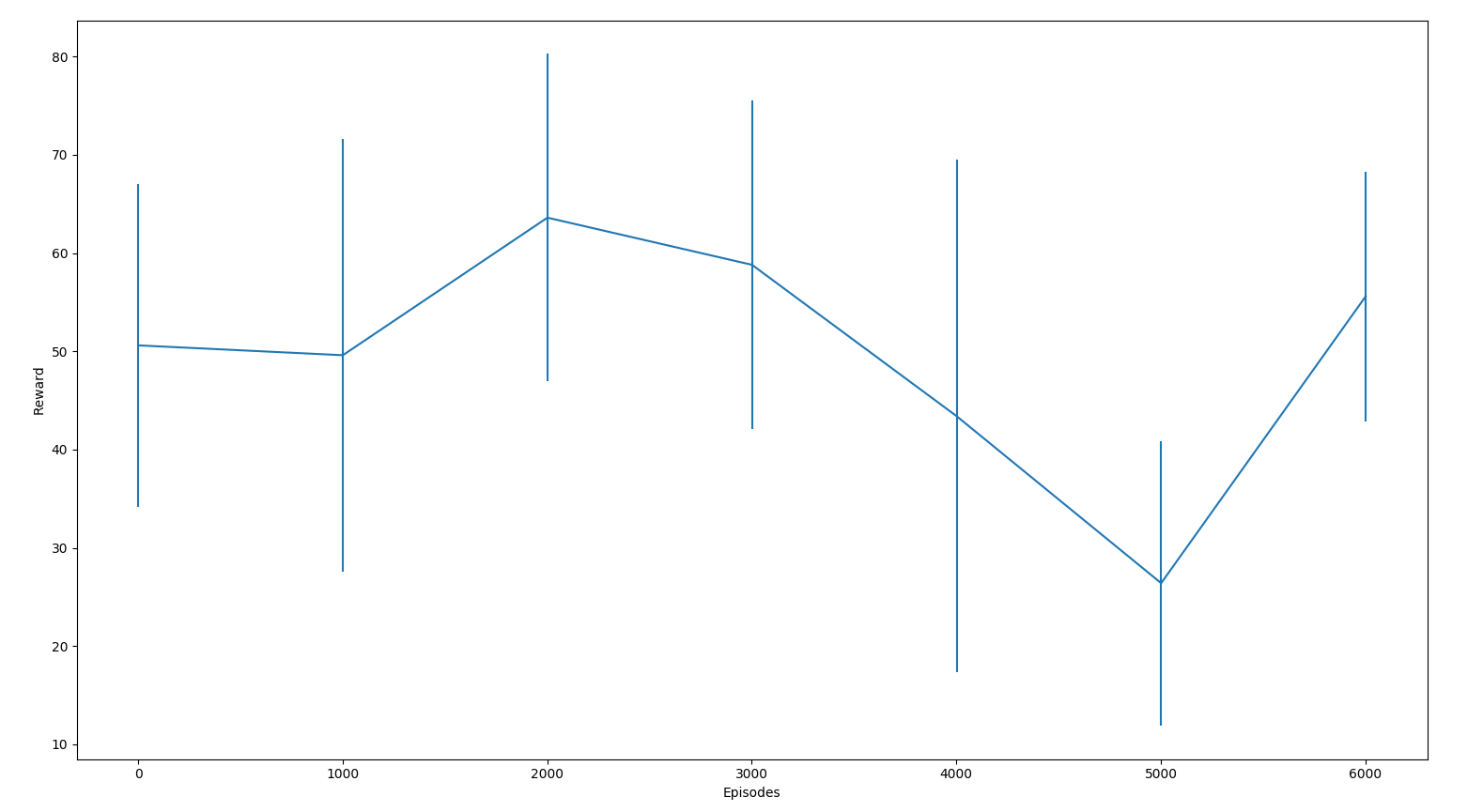}
  \caption{PlusFC from scratch vs. Baseline with transfer for the  DefeatZerglingsAndBanelings minigame}
  \label{DZAB}
\end{figure}

   \item BuildMarines: Here, the objective is to build marines by collecting minerals, which are then used to build supply depots and barracks, further used to build Marines. Our agent was only trained until 4k episodes( when it reached AWS storage maximum) up to which point the agent only learnt to move some SCVs to collect minerals. It never explored the other actions. Since it never builds any marines, rewards were zero. This is also the toughest minigame. Video of agent in action at 4k can be viewed \href{https://www.youtube.com/watch?v=a9VP04WNkpE&feature=youtu.be}{here}

   \item Simple64: It is the full game on a simple map where the agent is required to use all of the above scenarios simultaneously. Video of trained agent playing the composite game on Simple64 map can be viewed \href{https://www.youtube.com/watch?v=wmeprNS1S3Y&feature=youtu.be}{here}. To our best knowledge, there is no reported baseline performance on Simple64 map including at \href{http://starcraftgym.com/}{StarCraftGym}. Our results on this scenario are shown in \ref{simple64}.
   \begin{figure}[!h]
  \centering
  \includegraphics[width=0.6\textwidth]{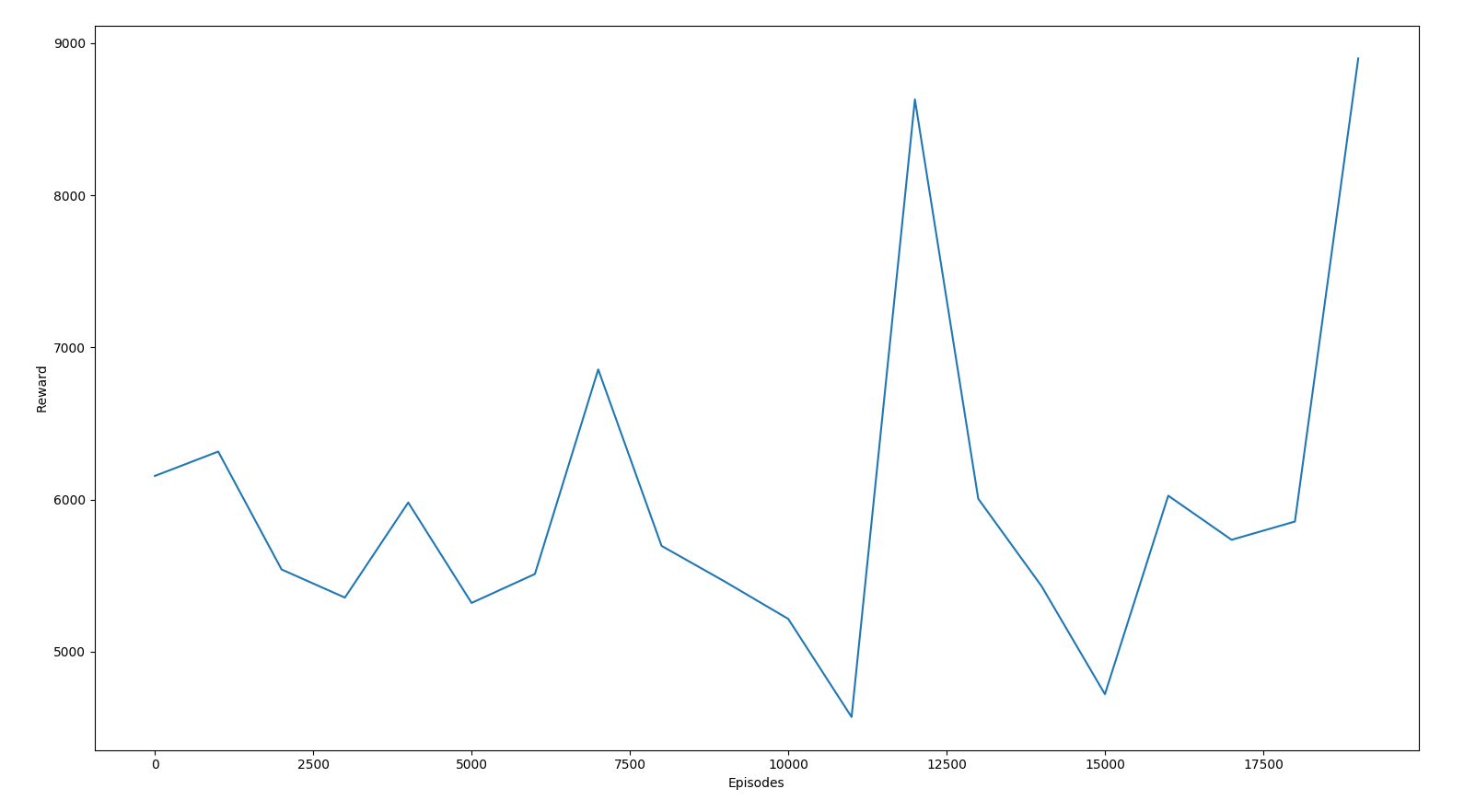}  
  \caption{Plot of scores on the composite Simple64 map for 18k episodes using Baseline network}
  \label{simple64}
\end{figure}
  
\end{enumerate}

\section{Discussion}
\label{discussion}

\subsection{Our Conclusions}
From this work, we can conclude the following:
\begin{enumerate}
  \item Given the complexity of the action space, the challenging part in this problem was to determine where on the map to take the action.
  \item The models often degraded which compelled us to prune by returning to a good checkpoint and retraining. 
  \item The more difficult maps requires additional complexity in the network to be able to learn effectively. AS a general rule, the PlusFC and PlusConv models perform better than Baseline model. 
  \item Transfer learning has been the most important breakthrough in this work. It drastically shortens training time and allows agents to explore new policies, not to mention that the transferred policy often does really well on newer maps. In order to solve maps of higher difficulty which are composites of easier minigames, transfer learning is surely the way to go. 
\end{enumerate}

\subsection{Future Work}

In the more complex scenarios, our agents converged on suboptimal strategies. Future work could investigate algorithms and network architectures that encourage further exploration. For example, Mnih, et. al. [9] recommend the addition of an entropy-based term to the value output for the A3C network.

Future work could investigate whether different network architectures can achieve greater performance. Our networks are all simple feed-forward, whereas a network including an LSTM layer would have memory. The insight of memory could allow the agent to map dependencies in time domain and learn better strategies. 

Finally, future work could take our methods and use them on more difficult challenges. For example, we could train on real Starcraft II maps instead of scenarios/minigames. We could also try transfer learning from the most complex scenario we tried (BuildMarines) to the Simple64 map or other full game maps.

\section*{References}

\small
[1] Volodymyr Mnih, Koray Kavukcuoglu, David Silver, Andrei A Rusu, Joel Veness, Marc G Bellemare, Alex Graves, Martin Riedmiller, Andreas K Fidjeland, Georg Ostrovski, et al. Human-level control through deep reinforcement learning. {\it Nature}, 518(7540):529–533, 2015.

[2] David Silver, Aja Huang, Chris J Maddison, Arthur Guez, Laurent Sifre, George Van Den Driessche, Julian Schrittwieser, Ioannis Antonoglou, Veda Panneershelvam, Marc Lanc- tot, et al. Mastering the game of Go with deep neural networks and tree search. {\it Nature}, 529 (7587):484–489, 2016.

[3] Charles Beattie, Joel Z Leibo, Denis Teplyashin, Tom Ward, Marcus Wainwright, Heinrich Kuttler, Andrew Lefrancq, Simon Green, Vıctor Valdes, Amir Sadik, et al. DeepMind Lab. {\it arXiv preprint arXiv}:1612.03801, 2016

[4] Sergey Levine, Peter Pastor, Alex Krizhevsky, Julian Ibarz, and Deirdre Quillen. Learning hand-eye coordination for robotic grasping with deep learning and large-scale data collection. The International Journal of Robotics Research, page 0278364917710318, 2016

[5]Andrei A Rusu, Matej Vecerik, Thomas Rothorl, Nicolas Heess, Razvan Pascanu, and Raia Hadsell. Sim-to-real robot learning from pixels with progressive nets. {\it arXiv preprint arXiv}:1610.04286, 2016.

[6]Vinyals, O., Ewalds, T., Bartunov, S., Georgiev, P., Vezhnevets, A. S., Yeo, M., … Tsing, R. (2017). StarCraft II: A New Challenge for Reinforcement Learning. \url{https://doi.org/https://deepmind.com/documents/110/sc2le.pdf}

[7] Steven Brown. Build a Sparse Reward PySC2 Agent. \url{https://itnext.io/build-a-sparse-reward-pysc2-agent-a44e94ba5255}

[8] Ilya Kostrikov. PyTorch implementation of Asynchronous Advantage Actor Critic (A3C). \url{https://github.com/ikostrikov/pytorch-a3c}

[9] Volodymyr Mnih, Adrià Puigdomènech Badia, Mehdi Mirza, Alex Graves, Timothy P. Lillicrap, Tim Harley, David Silver, Koray Kavukcuoglu. Asynchronous Methods for Deep Reinforcement Learning. \textit{ICML 2016}.

[10] Richard S. Sutton and Andrew G. Barto. 1998. Introduction to Reinforcement Learning (1st ed.). MIT Press, Cambridge, MA, USA.

[11] Arun Nair, Praveen Srinivasan, Sam Blackwell, Cagdas Alcicek, Rory Fearon, Alessandro De Maria, Vedavyas Panneershelvam, Mustafa Suleyman, Charles Beattie, Stig Petersen, Shane Legg, Volodymyr Mnih, Koray Kavukcuoglu, and David Silver. Massively parallel methods for deep reinforcement learning. In ICML Deep Learning Workshop. 2015.

[12] Xiaowei Hu. PySC2 agents. \url{https://github.com/xhujoy/pysc2-agents}

[13] Andrew G. Chang. Deep RL For Starcraft II. \url{http://cs229.stanford.edu/proj2017/final-reports/5234603.pdf}

[14] Jeffrey Barratt, Chuanbo Pan. Deep Imitation Learning for Playing Real Time Strategy Games. \url{http://cs229.stanford.edu/proj2017/final-reports/5244338.pdf}.

[15] Michal Certicky, David Churchill. The Current State of StarCraft AI Competitions and Bots. \url{http://agents.fel.cvut.cz/~certicky/files/publications/aiide17-certicky-churchill.pdf}.

[16] Niels Justesen, Sebastian Risi. Learning Macromanagement in StarCraft from Replays using Deep Learning. \url{https://arxiv.org/pdf/1707.03743.pdf}.

[17] Per-Arne Andersen, Morten Goodwin, Ole-Christoffer Granmo. Towards a Deep Reinforcement Learning Approach for Tower Line Wars. \url{https://link.springer.com/content/pdf/10.1007%2F978-3-319-71078-5_8.pdf}

\end{document}